\def\year{2020}\relax
\documentclass[letterpaper]{article} 
\usepackage{aaai20}  
\usepackage{times}  
\usepackage{helvet} 
\usepackage{courier}  
\usepackage[hyphens]{url}  
\usepackage{graphicx} 
\usepackage{graphicx}  
\usepackage{subcaption}
\usepackage{enumitem}
\usepackage{amssymb}
\usepackage{amsmath}
\usepackage{mathabx}
\usepackage{amsfonts} 
\usepackage{bbm}
\usepackage{multirow}
\usepackage{appendix}

\newcommand{\citet}[1]{\citeauthor{#1}~\shortcite{#1}}

\usepackage{array,tabularx}
\newenvironment{vardefs*}{\par\vspace{\abovedisplayskip}\noindent
	\tabularx{\columnwidth}{>{$}l<{$} @{${}:{}$} >{\raggedright\arraybackslash}X}}
{\endtabularx\par\vspace{\belowdisplayskip}}

\title{ORL: Reinforcement Learning Benchmarks for Online Stochastic Optimization Problems}

\author{\Large \textbf{Bharathan Balaji, Jordan Bell-Masterson, Enes Bilgin, Andreas Damianou, Pablo Moreno Garcia,} \\ \vspace{3mm}
	\Large \textbf{Arpit Jain, Runfei Luo, Alvaro Maggiar, Balakrishnan Narayanaswamy, Chun Ye} \\
	Amazon
}
 
\begin{document}

\maketitle

\begin{abstract}
Reinforcement Learning (RL) has achieved state-of-the-art results in domains such as robotics and games.  We build on this previous work by applying RL algorithms to a selection of canonical online stochastic optimization problems with a range of practical applications: Bin Packing, Newsvendor, and Vehicle Routing.  While there is a nascent literature that applies RL to these problems, there are no commonly accepted benchmarks which can be used to compare proposed approaches rigorously in terms of performance, scale, or generalizability. This paper aims to fill that gap.  For each problem we apply both standard approaches as well as newer RL algorithms and analyze results.  In each case, the performance of the trained RL policy is competitive with or superior to the corresponding baselines, while not requiring much in the way of domain knowledge. This highlights the potential of RL in real-world dynamic resource allocation problems. 
\end{abstract}

\section{Introduction}

Reinforcement learning (RL) has achieved state of the art results in gaming~\cite{silver2017mastering}, robotics~\cite{andrychowicz2018learning} and others. 
Our work relates to the growing literature of applying RL to optimization problems. \cite{bello2016neural} show RL techniques produce near optimal solutions for the traveling salesman (TSP) and knapsack problems. \cite{kool2018attention} use RL to solve TSP and its variants: vehicle routing, orienteering, and a stochastic variant of prize-collecting TSP. \cite{nazari2018reinforcement} solve both static and online versions of the vehicle routing problem. \cite{gijsbrechts2018can} apply RL to the dual sourcing inventory replenishment problem, and further demonstrate results on a real dataset. \cite{kong2018new} apply RL to online versions of the knapsack, secretary and adwords problems. \cite{oroojlooyjadid2017deep} apply RL to the beer game problem. \cite{lin2018efficient} use RL for fleet management of taxis on a real life dataset. 


Our contribution is to extend the existing RL literature to a set of dynamic resource allocation problems which parallel real-world problems. In particular, we present benchmarks for three classic problems: Bin Packing, Newsvendor and Vehicle Routing.  In each case, we show that trained policies from out-of-the-box RL algorithms with simple 2 layer neural networks are competitive with or superior to established approaches. 
We open source our code\footnote{https://github.com/awslabs/or-rl-benchmarks} and parameterize the complexity of the problems in order to encourage fair comparisons of algorithmic contributions.  Each environment is implemented with the OpenAI Gym interface~\cite{brockman2016openai} and integrated with the RLlib~\cite{liang2018rllib} library so researchers can replicate our results, test algorithms and tune hyperparameters. 

\section{Bin Packing}
\label{binpacking}

In the classic version of the bin packing problem, we are given items of different sizes and need to pack them into as few bins as possible. In the online stochastic version, items arrive one at a time with item sizes drawn from a fixed but unknown distribution. 
Many resource allocation problems in Operations Research and Computer Science face uncertain supply, and can be cast as variants of the online bin packing problem. In warehouse and transportation operations, variants of bin packing can be seen in: the order assignment problem (where we assign orders to fulfillment resources), the tote packing problem (where we fill items as they arrive into totes for shipment), and the trailer truck packing problem. In computing, bin packing problems arise in cloud computing scenarios, where virtual machines with varying memory and cpu requirements are allocated to servers with fixed capacity. 



\ifx 3d bin packing has many applications within Amazon operations from packing of picked items into totes to the packing of packages onto trailers. The need to use as few bins as possible directly translates into cost savings. \emph{Bharath} can you write about the VM
\fi

\subsection{Problem Formulation} \label{sec: bin_packing_prob_form}
We use a formulation of the bin packing problem similar to \citet{gupta2012online}. In the online stochastic bin packing problem, items arrive online, one in each time period $t$, with $t \in \{1, \ldots, T\}$. Items can be of different types $j \in \{1,...,J\}$. The size of type $j$ is $s_j$ and the probability that an item is of type $j$ is $p_j$. Without loss of generality, we assume item types are indexed in the increasing order of their size: $s_1 < s_2 < ... < s_J$. Upon arrival, the item needs to be packed into one of the bins, each with size $B$ (we assume that $s_J < B < \infty$). A packing is considered feasible if the total size of the items packed in each bin does not exceed the bin size. The task is to find a feasible packing that minimizes the number of bins used to pack all of the items that arrive within the time horizon. We assume the item sizes $s_j$ and bin size $B$ are integers. We assume the number of bins one can open is unlimited and denote the sum of item sizes in a bin $k$ as \emph{level} $h_{k}$. After $t$ items have been packed, we denote the number of bins at some level $h$ as $N_h(t)$, where $h \in \{1,...,B\}$. 

It can be shown that minimizing the number of non-empty bins is equivalent to minimizing the total waste (i.e. empty space) in the partially filled bins. In real applications (e.g. packing trucks, or virtual machines), there is a dollar-value cost associated with the consumption of these resources, so at any time horizon our objective is to minimize total waste $\sum_{t=0}^{T} W(t)$, where

\begin{equation}
\label{waste}
W(t) \triangleq \sum_{h=1}^{B-1} N_h(t)(B-h). 
\end{equation}

\noindent We use $W^{A}_{F}(t)$ to denote the total waste after step $t$ of algorithm $A$ when the input samples come from distribution $F$. For RL, we define the cumulative reward up to time step $t$ to be $W^{RL}_{F}(t)$. \citet{courcobetis1990stability} showed that any discrete distribution falls into one of three categories based on expected distribution $E[W^{OPT}_{F}(t)]$, where OPT is an offline optimal policy.
\begin{enumerate}[topsep=0pt,itemsep=-1ex,partopsep=1ex,parsep=1ex]
	\item Linear waste (LW): $E[W^{OPT}_{F}(t)] = \Theta(t)$, e.g. $B = 9$, two item types of size $\{2,3\}$ with probability $\{0.8, 0.2\}$ respectively.
	\item Perfectly Packable (PP):  $E[W^{OPT}_{F}(t)] = \Theta(\sqrt{t})$, e.g.  $B = 9$, two item types of size $\{2,3\}$ with probability $\{0.75, 0.25\}$ respectively.
	\item PP with bounded waste (BW): $E[W^{OPT}_{F}(t)] = \Theta(1)$, e.g. $B = 9$, two item types of size $\{2,3\}$ with probability $\{0.5, 0.5\}$ respectively.
\end{enumerate}
We will train an RL policy for each of the three distribution types and compare our policy to the appropriate baseline.

We formulate the bin packing problem as an MDP, where the state $S_t \in \mathcal{S}$ is the current item with size $s_j$ and the number of bins at each level is $N_h(t)$, where $h \in \{1,...,B\}$. The action $a$ is to pick a bin level which can fit the item.  Thus, the number of actions possible is $B$ with one action for each level and action 0 corresponds to opening a new bin. An episode defines the start and end of simulation. Initially, all the bins are empty. The reward $R_t$ is the negative of incremental waste as each item is put into a bin $s_j$. If the item is put into an existing bin, the incremental waste will reduce by item size. If the item is put into a new bin, the waste increases by the empty space left in the new bin. $T$ items need to be placed in the bins, after which the episode ends. We leave varying number of time steps for future work. We impose action masking for infeasible actions, such as picking a level for which bins do not exist yet, by outputting a probability for every action (regardless of feasibility) but multiplying the infeasible action probabilities by zero so they are never executed. 

\begin{figure*}
	\centering
	\begin{subfigure}[b]{0.33\textwidth}
		\centering
		\includegraphics[width=\textwidth]{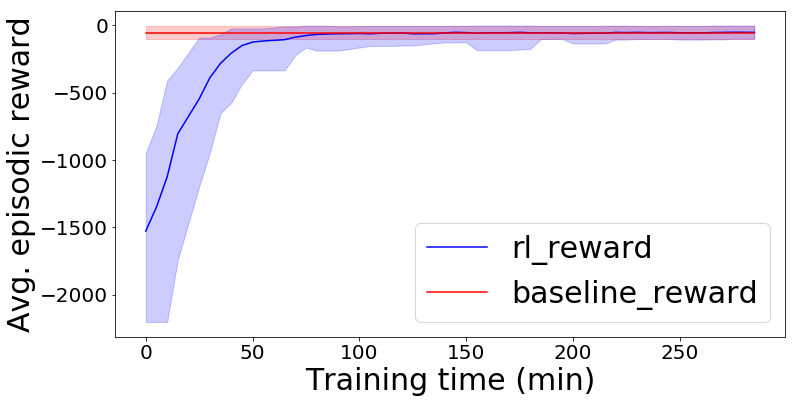}
		\caption{RL vs baseline for BW distribution}
		\label{fig:bin_packing_BW_dist1}
	\end{subfigure}
	\hfill
	\begin{subfigure}[b]{0.33\textwidth}
		\centering
		\includegraphics[width=\textwidth]{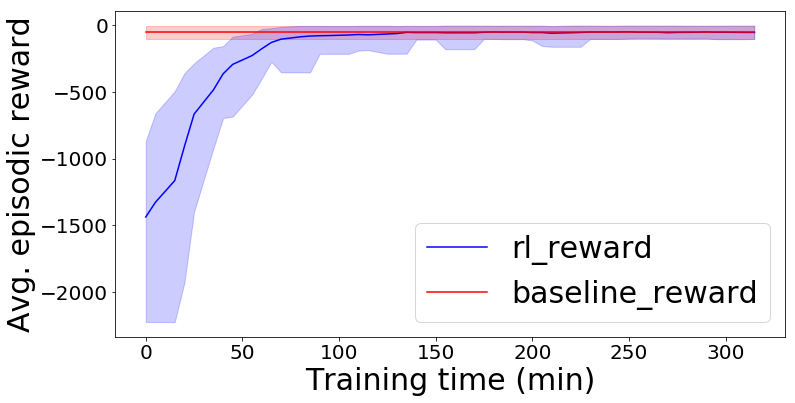}
		\caption{RL vs baseline for PP distribution}
		\label{fig:bin_packing_PP_dist1}
	\end{subfigure}
	\hfill
	\begin{subfigure}[b]{0.33\textwidth}
		\centering
		\includegraphics[width=\textwidth]{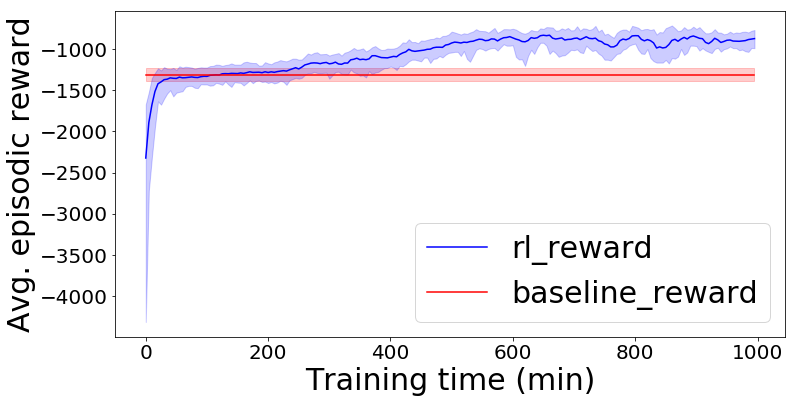}
		\caption{RL vs baseline for LW distribution}
		\label{fig:bin_packing_LW_dist1}
	\end{subfigure}
	\caption{Comparison of episodic rewards between RL and Best Fit baseline during training.}
	\label{fig:RLvsBF_binpacking}
\end{figure*}

\subsection{Related Work}
Bin packing is a well-studied problem in the operations research and computer science literature. 
The problem is already NP-hard in its basic form. As a result, many of the classical approaches to bin packing analyze the performance of approximation algorithms. We refer the readers to the survey \cite{Coffmanetal2013} for algorithmic approaches to classical bin packing and its generalizations.   

For online bin packing, a simple heuristic -- Best Fit -- is known to use at most $1.7$ times the optimal number of bins in the worst case \cite{Johnsonetal1974}. Best Fit places an item in a bin where, if the item were to be packed, would leave the least amount of space.
Another competitive heuristic is Sum of Squares (SS) heuristic \cite{csirik2006sum}. In particular, SS is proven to be asymtotically optimal (up to constants) as the episode length grows.

The simple heuristics described above are distribution agnostic. More sophisticated algorithms learn an empirical estimate of the item size distribution, leverage such distribution to solve a linear program, and use its dual to guide the online policy~\cite{IyengarSigman2004}. This approach has been used to solve online packing and covering problems \cite{GuptaMolinaro2014,AgrawalDevanur2015}.


\subsection{Baseline Algorithms}
We use the Sum of Squares (SS) heuristic and Best Fit (BF) as our baseline algorithms. 
When the $t$th item of size $s$ arrives, SS picks a bin of level $h^*$ that minimizes the value of the following sum-of-squares potential:

\begin{equation}
\sum_{h=1}^{B-1} (N_{h}(t))^2 \label{SOS eq}.
\end{equation}

\noindent It can be shown that minimizing (\ref{SOS eq}) is equivalent to:
\begin{equation}
h^* = \underset{h:N_h(t-1)>0 \ \text{and} \ h+s \leq B}{\arg\min} [N_{h+s}(t-1) - N_h(t-1)], \label{SOS eq1}.
\end{equation}

\noindent where, $N_0 = N_B = 0$. Intuitively, SS tries to equalize the number of bins at each level. Due its simplicity, we implemented (\ref{SOS eq1}) version of SS.

BF selects a bin at the highest level that can fit the item:
\begin{equation}
h^* = \underset{h:N_h(t-1)>0 \ \text{and} \ h+s \leq B}{\arg\max} h
\end{equation}

\subsection{Reinforcement Learning Algorithm}
We use the Proximal Policy Optimization (PPO) algorithm~\cite{schulman2017proximal}. 
We use a two-layer neural network with 256 hidden nodes each for both the actor and the critic. The input to both actor and critic network is the state, the output of the actor network is a vector giving the probabilities of taking any action in the action space, and the output of the critic network is predicted cumulative discounted reward from that state. During training, the agent explores the state space by sampling from the probability distribution of the actions generated by the policy network. We mask actions by reducing the probability of invalid actions to 0. We use a single machine with 4 GPUs and 32 CPUs for our experiments. 
We list all the hyperparameters used in Appendix \ref{appendix:bin_size_hp}.

\subsection{Results}

For each sample item size distribution (BW, PP, LW), we train the RL algorithm (PPO) and compare to the baseline algorithms (SS and BF).  We consider two variations, bin size of 9, with 100 items and distributions listed in section \ref{sec: bin_packing_prob_form}, and bin size of 100, with 1000 items and the following item size distribution:

\begin{enumerate}[topsep=0pt,itemsep=-1ex,partopsep=1ex,parsep=1ex]
	\item item sizes: $[1, 2, 3, 4, 5, 6, 7, 8, 9]$
	\item item probabilities for BW: \\ $[0.14, 0.10, 0.06, 0.13, 0.11, 0.13, 0.03, 0.11, 0.19]$
	\item item probabilities for PP: \\ $[0.06, 0.11, 0.11, 0.22, 0, 0.11, 0.06, 0, 0.33]$
	\item item probabilities for LW: $[0, 0, 0, 1/3, 0, 0, 0, 0, 2/3]$.
\end{enumerate}

%
%

Figure \ref{fig:RLvsBF_binpacking} plots the reward earned by the RL policy in training (blue) vs the Best Fit baseline (red) for bin size 100 and different item size distributions (BW, PP, and LW) as a function of training time (measured in minutes). The solid lines represent the mean reward of each policy, and the shaded bands represent the min/max rewards. By the end of training, RL either matches or outperforms the baseline policy for all three item size distributions. In particular, the reward gap between RL and baseline is the largest for LW distribution (which is expected, as both BF and SS are known to be sub-optimal for LW distribution). 


In Table \ref{table:bin_packing_RL_baseline_comp}, we inspect numerically the trained RL policy vs. baseline for bin size 100.  
Supporting what we observed in the initial figures, this table shows the final RL policy outperforms or matches the baseline for each distribution.

We test generalization of the RL policy by evaluating the trained policy with a different item distribution than the one it was trained on. For PP and BW distributions, the trained policies translate well. Both the PP and BW policies perform as well as the baseline solutions for the LW distribution. The policy trained on the LW distribution generalizes reasonably well but does not do as well as the baseline solutions in the BW and PP distributions. We did observe overfitting if we pick model iterations from much later in training. We leave the study of overfitting and generalization across distributions as future work. A note on scaling: the training time for bin size 100 is about 3x, 4x and 10x more than bin size 9 for PP, BW and LW respectively. The bin size 9 results can be found in the supplementary material.

\begin{table}[h!]
	\resizebox{\columnwidth}{!}
	{%
		\begin{tabular}{|c|cc|cc|cc|}
			\hline
			\multicolumn{1}{|l|}{\multirow{2}{*}{Algorithm}}  & \multicolumn{2}{c|}{Perfect Pack} & \multicolumn{2}{c|}{Bounded Waste}	& \multicolumn{2}{c|}{Linear Waste} \\
			\multicolumn{1}{|l|}{} 											&   $\mu$            &        $\sigma$		& 	$\mu$    			& 	$\sigma$     		 &     $\mu$        &      $\sigma$          \\ \hline
			RL with PP  										  & -49.0     		   &          29.5        	&  	-48.0 	  			  &   29.5	 				 &     -1358       &      44.2     \\ \hline
			RL with BW  									  & -47.6     		   &          29.3        	&  	-53.9 	  			  &   26.4	 				 &     -1368       &      48.0     \\ \hline
			RL with LW 										 &  -258.6     		  &         69.3           &  	-143.9 	  			&   84.9	 			   &     -880.2      &      43     \\ \hline
			SS  											   &  -56.54     	    &         28.9           &    -56.61 	  			&   30.2	 			   &     -2091      &      92     \\ \hline
			Best Fit  															  &  -52.01     	   &          29.5           &  	-51.4 	  			&   28.9	 			   &     -1314     &      53     \\ \hline
		\end{tabular}
	}
	\caption{RL and baseline solution comparison for bin packing. Mean and standard deviations are across 100 episodes.}
	\label{table:bin_packing_RL_baseline_comp}
\end{table}

\begin{figure}[h!]
	\centering
	\includegraphics[width=1\linewidth]{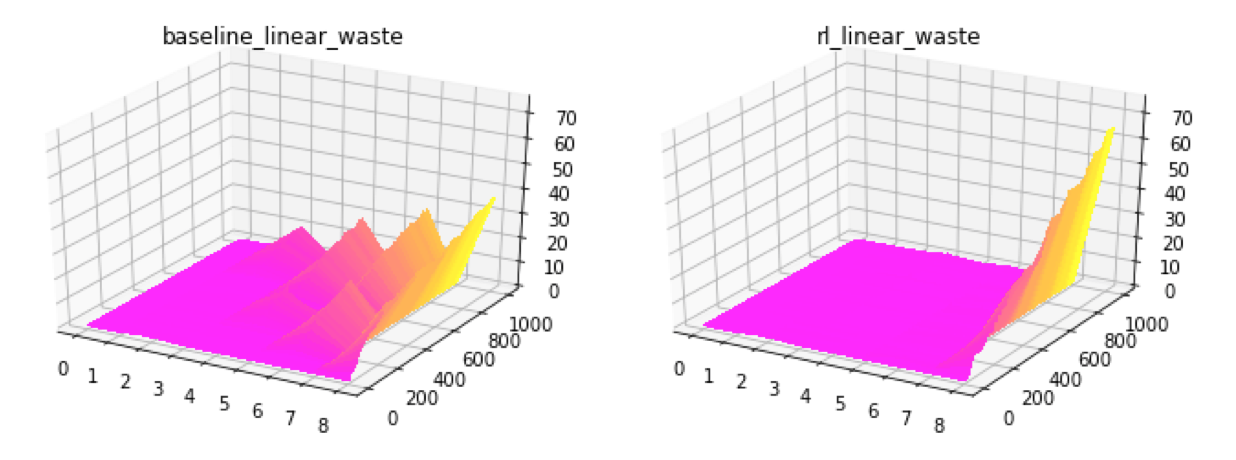}
	\caption{RL vs baseline solution for LW distribution}
	\label{fig:bin_packing_LW_dist}
\end{figure}

Finally, we inspect the relative structure of the policies to ensure that RL is learning a sensible solution.  In particular, we plot the state variable values as a function of the number of steps in an episode. Intuitively, the integral of these plots represents the waste, which we want to minimize.  An optimal policy should show a (relatively) flat surface. We use bin size of 9 for this analysis for ease of manual inspection and study the linear waste distribution that highlights the difference between the Sum of Squares baseline and RL distinctly. From Figure \ref{fig:bin_packing_LW_dist}, we see that the baseline policy leaves more open bins at a lower fullness, whereas RL only leaves open bins at level 8 (which cannot be closed once they reach that level). This indicates that the learned RL polocy is reasonable. For other distributions, the graphs for both the baseline and RL policy look similar to each other.



\section{Multi-Period Newsvendor with Lead Times}
\label{sec:newsvendor}

The Newsvendor problem (see e.g. \citet{zipkin2000foundations}) is a seminal problem in inventory management wherein we must decide on an ordering decision (how much of an item to purchase from a supplier) to cover a single period of uncertain demand. The objective is to trade-off the various costs incurred and revenues achieved during the period, usually consisting of sales revenue, purchasing and holding costs, loss of goodwill in the case of missed sales, and the terminal salvage value of unsold items.

In practice, decisions are rarely isolated to a single period, and they are repeatedly and periodically taken and thus have a downstream impact. \emph{This makes the problem non-trivial and has no known optimal solution}, as compared to the single-period Newsvendor which has a known optimal solution when the demand distribution is known. Additionally, purchased units do not, in general, arrive quasi-instantaneously, but rather after a few periods of transit from the vendor to their final destination, known as the lead time. The presence of lead times further complicates the problem. Solving the multi-period newsvendor problem with lead times and lost sales is a notoriously difficult problem \cite{zipkin2008old}. It requires keeping track of orders placed in different periods, leading to what is known as the curse of dimensionality, rendering any exact solution impractical even for small lead times of 2 and 3 periods, and outright infeasible at higher dimensions. As a result, the problem forms a good test-bed for RL algorithms given that the observation of rewards is delayed by the lead time and that it can be formulated as a Markov Decision Problem. A number of heuristics have been developed for the lost sales problem, often based on order-up-to level policies for the equivalent model with backlogged demand. Comparisons in the performance of these two policies have been studied \cite{janakiraman2007comparison}, and it has been shown that order-up-to policies are asymptotically optimal \cite{huh2009asymptotic}, thus making for good benchmark policies.

\begin{figure}
	\centering
	\includegraphics[width=0.45\textwidth]{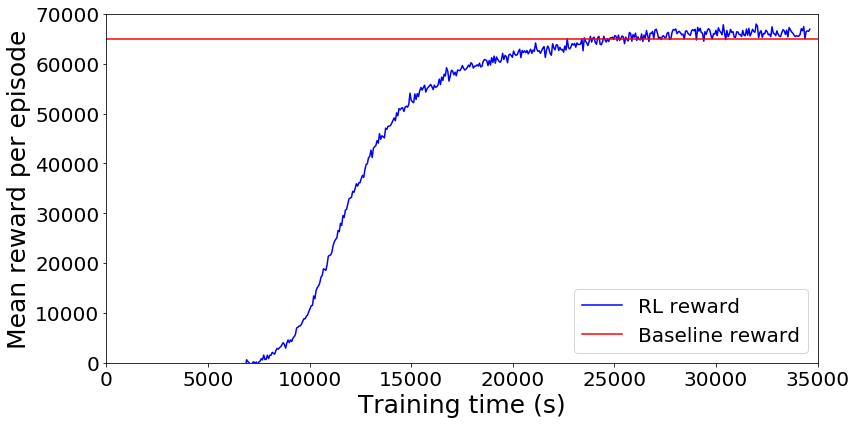}
	\caption{RL training reward in the multi-period newsvendor problem with Poisson demand and vendor lead period of 5.}
	\label{fig:newsvendor_TRPO}
\end{figure}

\begin{figure*}
	\centering
	\begin{subfigure}[b]{0.23\textwidth}
		\centering
		\includegraphics[width=\textwidth]{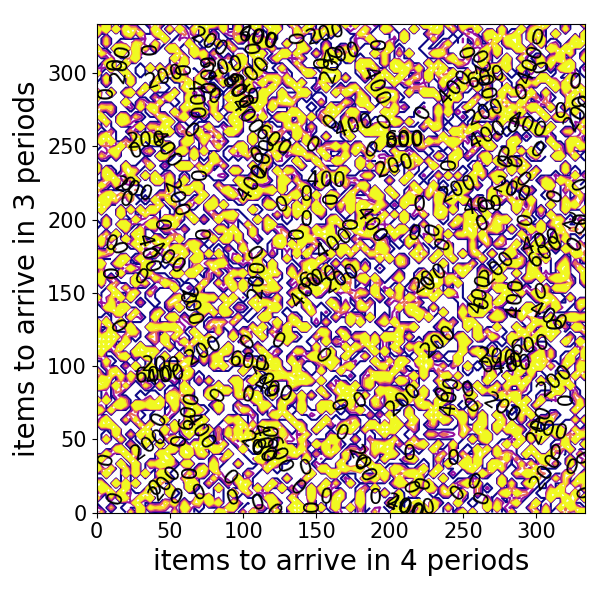}
		\caption{PPO Checkpoint 50}
		\label{fig:nv_ckpt_50}
	\end{subfigure}
	\hfill
	\begin{subfigure}[b]{0.23\textwidth}
		\centering
		\includegraphics[width=\textwidth]{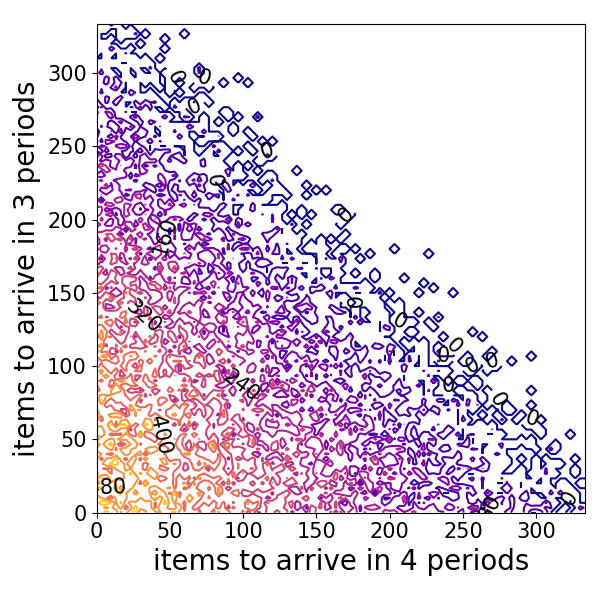}
		\caption{PPO Checkpoint 1000}
		\label{fig:nv_ckpt_1000}
	\end{subfigure}
	\hfill
	\begin{subfigure}[b]{0.23\textwidth}
		\centering
		\includegraphics[width=\textwidth]{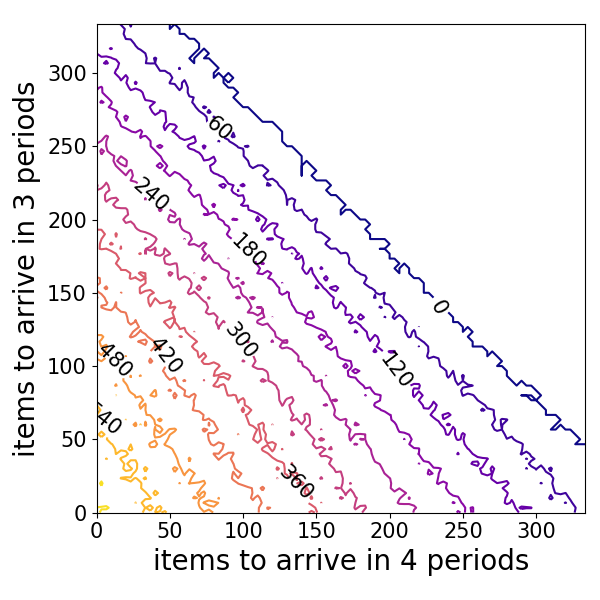}
		\caption{PPO Checkpoint 1500}
		\label{fig:nv_ckpt_1500}
	\end{subfigure}
	\hfill
	\begin{subfigure}[b]{0.23\textwidth}
		\centering
		\includegraphics[width=\textwidth]{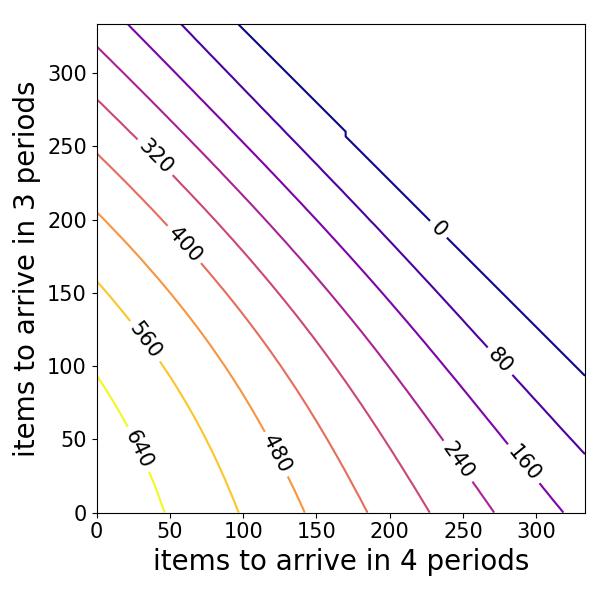}
		\caption{PPO Checkpoint 10050}
		\label{fig:nv_ckpt_10050}
	\end{subfigure}
	\begin{subfigure}[b]{0.035\textwidth}
		\centering
		\includegraphics[width=\textwidth]{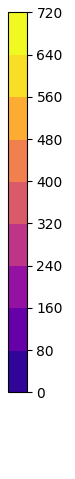}
		\label{fig:nv_ckpt_10050}
	\end{subfigure}
	\caption{The Newsvendor policy graphs show the RL-learned policy for the quantity we will buy, as a function of how much inventory we have already ordered. The axes show the inventory to arrive in three and four periods respectively and the contour lines show the number of items bought by the RL policy. The agent policy improves over the training period. In the final policy, If we have already ordered a lot of inventory, this graph shows we will order less at this timestep.}
	\label{fig:newsvendor_policy}
\end{figure*}

\subsection{Problem formulation}
We consider the stationary, single-product, multi-period dynamic inventory management problem with vendor lead time (VLT) and stochastic demand. Here, the VLT $l$  refers to the number of time steps between the placement  and receipt of an order. The demand $D$ is assumed to be stationary and Poisson distributed with mean $\mu$. Items are purchased at a cost $c$ and sold at a price $p$, and incur a penalty for lost sales $k$ for each unit of unmet demand while any unit left over at the end of a period incurs a holding cost $h$. Items do not perish. A discount factor $\gamma$ is used.  No terminal value is awarded for the inventory state at end of episode.

The problem is formulated as a Markov Decision Process: 
\begin{description}[style=unboxed,leftmargin=0cm]
	\item[State:] The state $S$ of the problem is given by
	\begin{align*}
	S=(p,c,h,k,\mu,x_0,\ldots,x_{l-1})
	\end{align*}
	where $x_0$ is the on-hand inventory, $x_1$ the units to be received one period hence, and so on.
	\item[Action:] In each period the state of the system is observed and an action $A=q$ is taken, consisting of the size of the order placed and to arrive $l$ time periods later.
	\item[Reward:] We first incur the purchasing cost corresponding to the procured units given the action $a$.
	A realization $d$ of the demand $D$ (Poisson distributed with mean $\mu$) is then observed, and demand is satisfied as much as is possible given on-hand levels. Missed sales incur a loss of goodwill $k$ per unit, while leftover units incur a holding cost $h$:
	\begin{align*}
	R &= p\min(x_0,d) - c a - h (x_0 - d)^+ - k (d - x_0)^+.
	\end{align*}
	where $(x)^+=\max(x,0)$.
	\item[Transition:] The state of the system $S$ is then updated to $S_+$ by moving all pipeline units downstream and incorporating the newly purchased units:
	\begin{align*}
	S_+ &= (p,c,h,k,\mu,(x_0 - d)^+ +x_1,x_2,\ldots,x_{l-1},a).
	\end{align*}
We do not impose action masking because there is no infeasible action in the pre-specified, positive continuous action space.  We convert the continuous buy quantity to integer by post-process rounding. 
\end{description}

\subsection{Related work}
Data-centric approaches \cite{rudin2014big} and reinforcement learning approachse \cite{oroojlooyjadid2016applying} have recently been suggested for the newsvendor problem. These have so far still remained focused on the single period problem and often trying to learn some of the inputs, such as demand. A few other papers have considered Reinforcement Learning in the context of inventory management, such as \cite{gijsbrechts2018can}, where a dual sourcing problem is tackled using RL.

\subsection{Baseline Algorithm} 

As noted in the beginning of this section, it is impractical or even infeasible to solve the multi-stage newvendor problem exactly. However, it is possible to use heuristics that provide good approximations to the optimal solution. In particular, a way to tackle the problem is to approximate it by its backlogging counterpart, where orders are not lost if unsatisfied, for which a closed form solution of the optimal policy exists in the form of an order-up-to policy characterized by the following critical ratio:
\begin{align*}
CR = \frac{p-\gamma c + k}{p-\gamma c + k + h}.
\end{align*}
As a result, letting $z^* = F_l^{-1}(CR)$, where $F_l$ is the cumulative distribution  function of the $l$ period demand, the policy is given by:
\begin{align*}
a&= \left(z^* - \sum_{i=0}^{l-1} x_i\right)^+.
\end{align*}

\subsection{Reinforcement Learning Algorithm} 

We use Proximal Policy Optimization (PPO) \cite{schulman2017proximal} as implemented in the RLlib package \cite{rllab}, where the policy is represented by a neural network. We use a single machine with 4 GPUs and 32 CPUs for our experiments. We use a fully-connected neural network with hidden layers of size (64,32) and the hyperparameters presented in Appendix \ref{appendix:newsvendor_hp}. 

\subsection{Results}

We present the results obtained using a VLT of 5 and time horizon of 40. The economic parameters were chosen so that $p,c\in[0,100]$, $h\in[0,5]$ and $k\in[0,10]$, while the demand mean $\mu$ was such that $\mu\in[0,200]$.

We sampled problem parameters as follows: $p\sim U[0,100]$, $c\sim U[0,p]$, $h\sim U[0,\min(c, 5)]$, $k\sim U[0,10]$ and $\mu\sim U[0,200]$ for the economic and demand parameters; where $U[a,b]$ denotes a uniformly random variable between $a$ and $b$.  The initial state was simply set to be $\mathbf{0}$.
Figure~\ref{fig:newsvendor_TRPO} compares the results obtained by the RL algorithm to the baseline. The RL solution eventually beats the benchmark.

While solving this problem numerically is intractable, the optimal inventory policy structures are well known. It is thus of interest to check whether their properties are being learned by the RL algorithm. Given the dimension of the problem, we cannot observe the entire policy, but can investigate slices thereof. We thus fix price, cost, holding cost, penalty for lost sale and mean demand to 50, 25, 0.5, 5 and 100, respectively, and plot the optimal policy in the space $(0,0,0,x_3,x_4)$ in Figure~\ref{fig:newsvendor_policy}, where $x_3$ and $x_4$ stand for the inventory to arrive in three and four periods respectively. The figure shows contour curves of the buying quantity as a function of the inventory state. Intuitively, a good policy will buy less if we already have inventory on-hand (or in the pipeline). Visually, this should look like a smooth, decreasing frontier.  We observe that the algorithm is learning this desired policy structure over the training period and we can start to observe monotonicity of the policy along most directions.


\section{Vehicle Routing Problem}
\label{sec_vrp_intro}

One of the most widely studied problems in combinatorial optimization is the traveling salesman problem (TSP), which involves finding the shortest route that visits each node in a graph exactly once and returns to the starting node. TSP is an NP-hard problem and has a variety of practical applications from logistics to DNA sequencing. The vehicle routing problem (VRP) is a generalization of TSP where one or more vehicles are expected to visit the nodes in a graph, for example to satisfy customer demand. VRP is also a well-studied topic and has several applications, especially in supply chain and logistics. 
An important extension of VRP is where some of the information about the graph is revealed over time, such as demand at each node and travel time. This class of VRP is called dynamic VRP (DVRP, also known as real-time or online VRP). Stochastic VRP (SVRP) is where one or more problem parameters are stochastic with some known probability distributions (as opposed to arbitrary or adversarial distributions). In many real-life applications, the relevant VRP is both stochastic and dynamic (SDVRP), which is also focus of this work. We formulate a variant of SVRP and compare solution approaches from the Operations Research (OR) and Reinforcement Learning (RL) literature.

\subsection{Problem Formulation}
\label{sec_vrp_pf}

We consider a VRP variant that is of an on-demand delivery driver. Orders arrive on the phone app of the driver in a dynamic manner throughout the problem horizon. Each order has a delivery charge (reward) known to the driver at the time of order creation, and it is assigned to a pickup location (e.g. restaurant) in the city. “City” here refers to the Euclidean space in a grid map where the VRP environment is created. The city consists of mutually exclusive zones that generate orders at different rates. At each time step, an order generated with a constant probability and assigned to a zone (i.e. $p_1=0.5,p_2=0.3,p_3=0.1,p_4=0.1$ for zones 1,…,4). Orders have rewards that come from zone-specific truncated normal distribution with different ranges (i.e. with minimum and maximum dollar values of [8,12], [5,8], [2,5] and [1,3] for each zone, respectively). Orders have delivery time windows, which is within 60 minutes from the creation of the order. The driver has to accept an order and pick up the package from a given location prior to delivery. Orders that are not accepted disappear probabilistically (i.e. with a time-out probability of 0.15 per time step) and assumed to be taken by some other competitor driver in the city. The vehicle has a capacity limit of 4 orders on the vehicle, but the driver can accept unlimited orders and plan the route accordingly. The driver incurs a cost per time step and unit distance traveled (0.1 for both), representing the money value of time and travel costs. The driver’s goal is to maximize the total net reward over an episode of 1000 time steps. This version of VRP is known as stochastic and dynamic capacitated vehicle routing problem with pickup and delivery, time windows and service guarantee.  We choose this particular variant, which is less studied in the literature, because it more closely resembles real-world instances of the problem and gives us higher confidence that RL can generalize beyond our toy setup.

\begin{description}[style=unboxed,leftmargin=0cm]
	\item [State:] 
	We include pickup location $p_t$, driver info, and order info. Driver info contains the driver's position $h_t$ and the capacity left $c_t$. Order info contains the orders' location $l_t$, status $w_t$ (open, accepted, picked up or delivered/inactive), the time elapsed since each order's generation $e_t$ and the corresponding dollar value $v_t$.  Thus, the state is  $S_t = (p_t, h_t, c_t, l_t, w_t, e_t, v_t)$.
	\item [Action] 
	The agent chooses an action $A_t$ from five options -- accept the open order $i \in P$, pick up the accepted order $i \in A$, pick up the accepted order $i \in A$, the pickup location $j \in R$, or wait and stay unmoved. 
	\item [Reward:] The reward $R_t$ is the total value of all delivered orders $f_t$ minus the cost $q_t$. $f_t$ is divided into $3$ equal parts for reward shaping: when the order gets accepted, picked up, and delivered respectively. Thus we have:
	\begin{equation*}
	R_t = \frac{1}{3}\Big(\mathbbm{1}_{accepted} + \mathbbm{1}_{picked-up} + \mathbbm{1}_{delievered} \Big) f_t - q_t,
	\end{equation*}
	where $q_t = (q_{time} + q_{move} + q_{failure})$. $q_{time}$ is the time cost, $q_{move}$ is the moving cost (per time step). $q_{failure}$ is a large penalty ($50$) if the agent accepts an order but fails to deliver within the promised time. 
	
\end{description}

The vehicle's capacity remains unchanged if an order is accepted but not picked up. In effect, this grants the agent the flexibility to accept more orders than available capacity, which can be picked up later when space allows. The action of heading to a specific pickup location enables the agent to learn to stay near popular pick up locations. 
We impose action masking during the policy training. The agent cannot perform the following invalid actions: \textit{(i)} pick up an order when its remaining capacity is $0$; \textit{(ii)} pick up an order that is not yet accepted; \textit{(iii)} deliver an order that is not in transit. 

\subsection{Related Work} \label{sec_vrp_related_work}
There is a substantial literature on VRP~\cite{EKSIOGLU20091472}. The closest VRP variant to the problem considered in this paper is the Pickup and Delivery Problem with Time Windows (PDPTW) \cite{Cordeau2008}, which has some additional complexities over vanilla VRP. Due to such complexities, there are fewer exact solution approaches \cite{LuDessouky2004,MAHMOUDI201619}, and the majority of the literature focuses on heuristics. When the problem is also stochastic and dynamic, exact solution methods become intractable except for very specific problem settings. In such cases, anticipatory algorithms that simulate sample future scenarios and merge solutions to those samples are a common choice \cite{Ulrike2016}.

Reinforcement Learning (RL) methods have been successfully used for solving the Traveling Salesman Probelm (TSP). \citet{bello2016neural} employ a pointer network to optimize the policy, and train an actor-critic algorithm with the negative tour length as the reward signal. \citet{khalil2017learning} develop a single model based on graph embeddings. They use the DQN algorithm to train a greedy policy and graph embedding network simultaneously.  For VRP, \citet{kool2018attention} utilize the transformer neural network architecture to develop a model fully based on attention layers. Their proposed model is trained by policy gradients with a greedy baseline, and evaluated on both standard Capacitated VRP (CVRP) and Split Deliverry VRP (SDVRP). \citet{nazari2018reinforcement} further improve the algorithm using embedded inputs and allow the customers and their demands to be stochastic. 


\subsection{Baseline Algorithm}
We modify the classical three-index Mixed Integer Programming (MIP) formulation \cite{RopkeCordeau2009,FURTADO2017334}. This deterministic MIP is solved for the available orders in the environment. It is further resolved when a new order arrives, if one of the existing orders expires, or when all of the actions are executed. When we solve the MIP, the orders that had been already accepted or were in transit are modeled as starting conditions. The details of our MIP model is in Appendix \ref{appendix:vrp_mip}. We leave anticipatory models to future work.


\ifx
\subsubsection*{Sets}
\begin{vardefs*}
	V & Current vehicle location, $V=\{0\}$ \\
	P & Pickup nodes (copies of the restaurant nodes, associated with the orders that are not in transit)  \\
	D & Delivery nodes representing the orders that are not in transit, $D = \{j | j= i + n, i \in P, n=|P| \}$  \\
	A & Nodes representing the orders that are accepted by the driver; $A \subset D$ \\
	T & Delivery nodes representing the orders that are in transit  \\
	R & Nodes representing the restaurants, used for final return) \\
	N & Set of all nodes in the graph, $N = V \cup P \cup D  \cup T \cup R $\\
	E & Set of all edges, $E=\{(i, j),  \forall i, j \in N\}$
\end{vardefs*}

\subsubsection*{Decision variables}
\begin{vardefs*}
	x_{ij} & Binary variable, 1 if the vehicle uses the arc from node $i$ to $j$, 0 otherwise; $i, j \in N$ \\
	y_{i}  & Binary variable, 1 if the order $i$ is accepted, 0 otherwise; $i \in P$\\
	Q_{i} & Auxiliary variable to track the capacity usage as of node  $i$; $i \in N$ \\ 
	B_{i} & Auxiliary variable to track the time as of node  $i$; $i \in N$
\end{vardefs*}

\subsubsection*{Parameters}
\begin{vardefs*}
	n & Number of orders available to pick up, $n = |P|$ \\ 
	c_{ij} & Symmetric Manhattan distance (in miles) matrix between node $i$ and $j$, $(i, j) \in E$ \\
	q_i & Supply (demand) at node $i$, $q_0 = |T|; q_i = 1, \forall i \in P;  q_i = -1, \forall i \in D \cup T; q_i = 0 \in R$  \\ 
	l_i & Remaining time to deliver order $i$, $i \in D \cup T$ \\ 
	m & Travel cost per mile \\
	r_i & Revenue for order associated with pick up node $i$, $i \in P$  \\
	U & Vehicle capacity  \\
	M & A very big number  \\ 
	t & Time to travel one mile  \\
	d & A constant positive service time spent on accept, pickup, delivery
\end{vardefs*}

\subsubsection*{Model}
\begin{equation}
\begin{array}{rrclcl}
& \max_{x, y, Q, B} & \multicolumn{3}{l}{ \sum_{i \in P} r_i y_i - m \sum_{(i,j) \in E} c_{ij} x_{ij}  } \\  
& \textrm{s.t.} \qquad  \sum_{j \in N} x_{ij}   &=& y_i & \forall i \in P \\   
& \sum_{j \in N} x_{ij} - \sum_{j \in N} x_{i+n,j}  & = & 0 & \forall i \in P  \\     
& y_i & =& 1 & \forall i \in A \\  
& \sum_{j \in N} x_{ij}   &=& 1 & \forall i \in V \cup T \\  
& \sum_{i \in N \setminus R } \sum_{j \in R }  x_{ij}   &=& 1  &\\     
& \sum_{j \in N \setminus R } x_{ji} - \sum_{j \in N} x_{ij}  & = & 0 & \forall i \in P \cup D \cup T  \\     
& Q_i + q_j - M (1-x_{ij} ) &\leq& Q_j & \forall i,j \in N \\  
& \max{(0, q_i)}    &\leq& Q_i & \forall i \in N \\  
& \min{(U, U+q_i)}    &\geq& Q_i & \forall i \in N \\  
& B_i + d + c_{ij} t -  M (1-x_{ij} )  &\leq& B_j   & \forall i,j \in N \\  
& B_i +  c_{i, i+n} t -  M (1- y_i )  &\leq& B_{i+n}  & \forall i \in P \\ 
&  d \sum_{i \in P \setminus A}  y_i  & = & B_0  \\ 
& B_i  &\leq& l_i  & \forall i \in D \cup T \\ 
& x_{ij}, y_i &\in& \{0, 1\} & \forall i,j \in N  \\ 
\end{array}
\end{equation}
\fi

\begin{figure*}[h!]
	\centering
	\begin{subfigure}[b]{0.45\textwidth}
		\centering
		\includegraphics[width=1\linewidth]{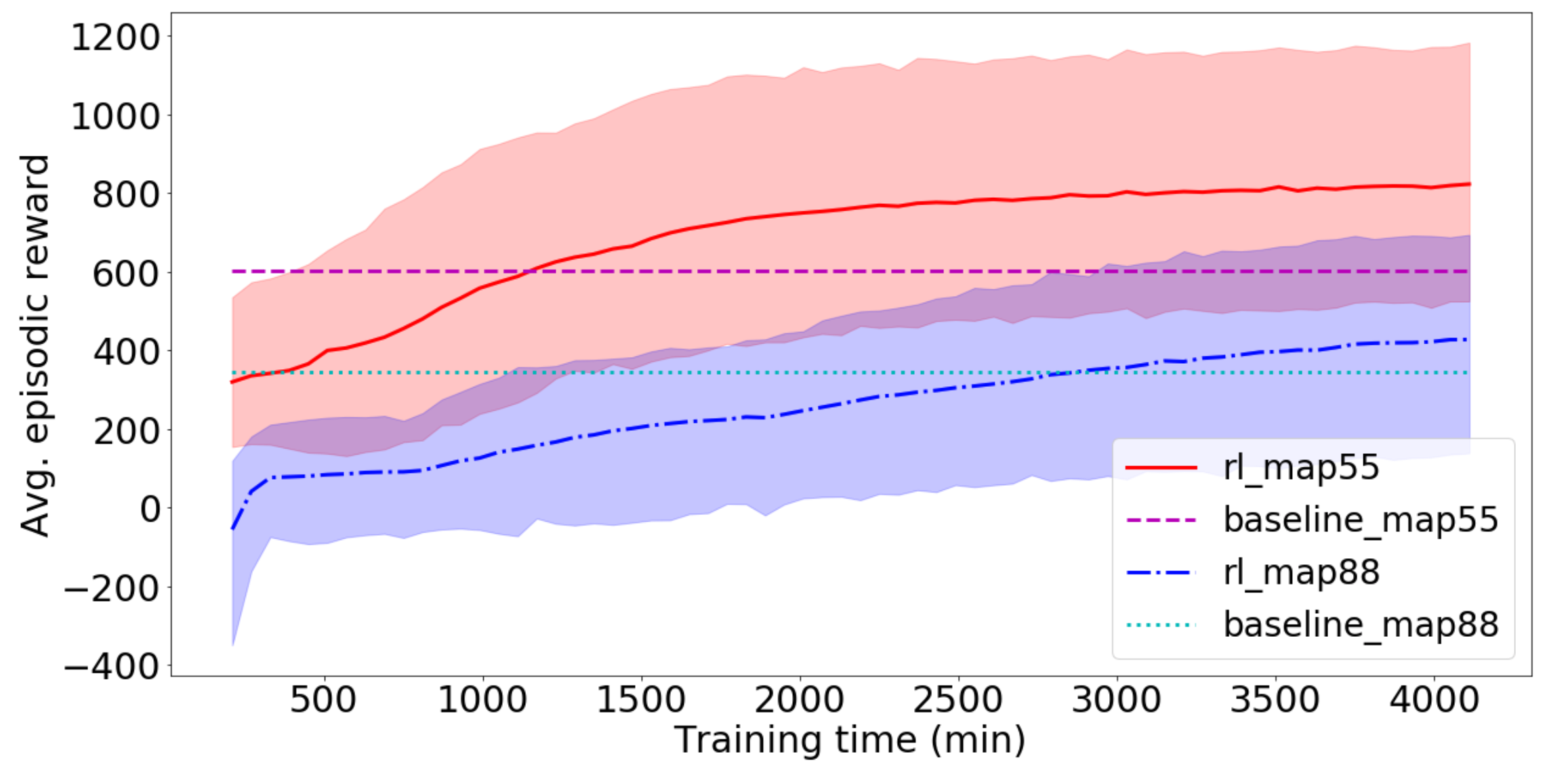}
		\caption{RL vs baseline solution for VRP with 3 pick-up locations, 5 orders and map sizes $5 \times 5$ and $8 \times 8$ }
		\label{fig:vrp_mapsize}
	\end{subfigure}
	\hspace{2em}
	\begin{subfigure}[b]{0.45\textwidth}
		\centering
		\includegraphics[width=1\linewidth]{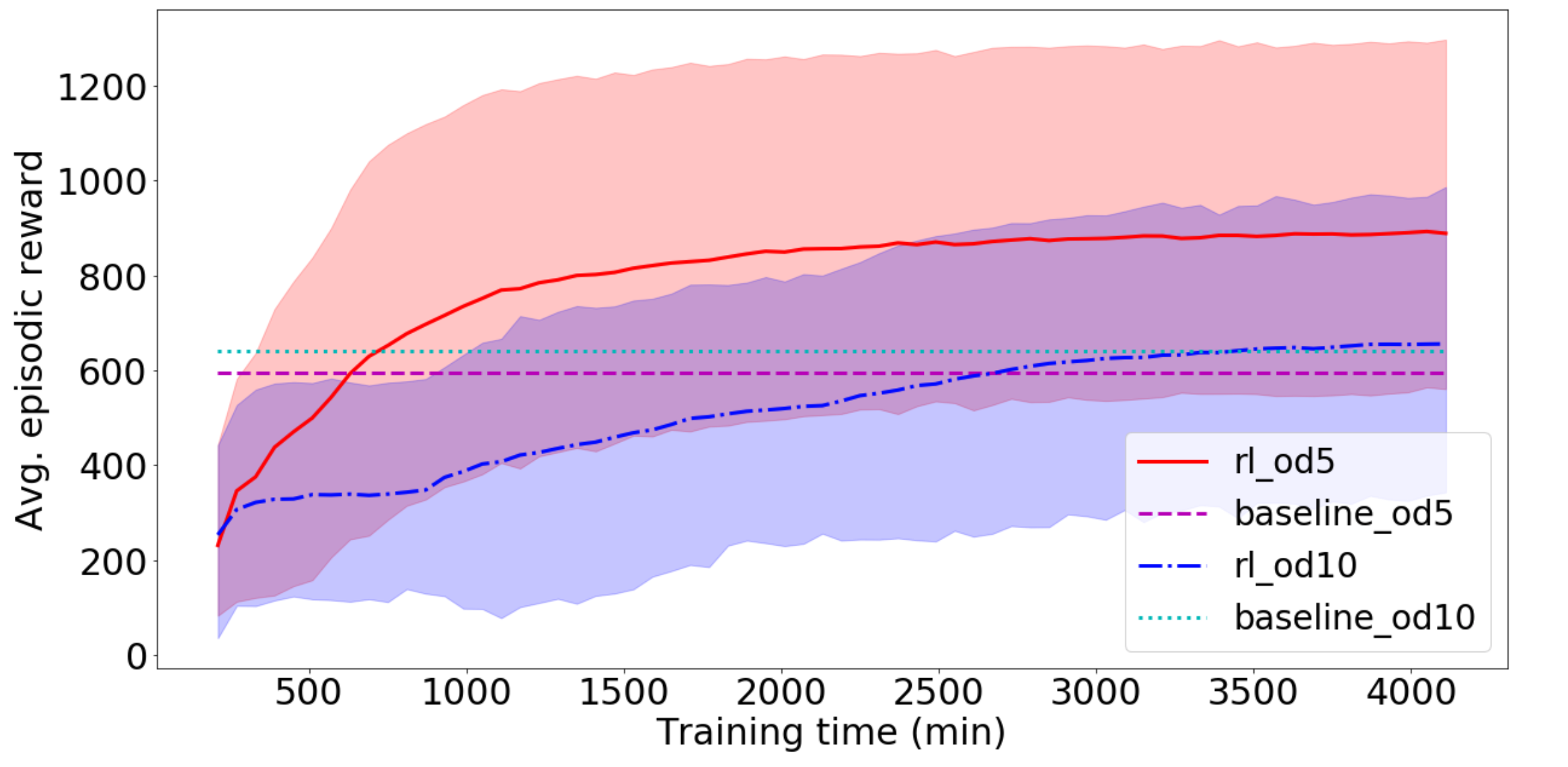}
		\caption{RL vs baseline solution for VRP with 2 pick-up locations, map $5 \times 5$, and number of orders 5 and 10. }
		\label{fig:vrp_ordernumber}
	\end{subfigure}
	\label{fig:vrp}
	\caption{RL vs baseline during policy training process.}
\end{figure*}

\subsection{Reinforcement Learning Algorithm}
\label{subsec:vrp_rl}

To train the policy, we apply the APE-X DQN algorithm~\cite{horgan2018distributed} due to its ability to scale by generating more experience replays and picking from them in a distributed prioritized fashion\footnote{We also applied PPO with default hyperparameters provided in RLLib. The reward increased much slower than that of APEX-DQN and was not able to beat the baseline after 1 day of training.}. We use a two-layer neural network with 512 hidden units each. We list all hyperparameters in Appendix \ref{appendix:vrp_hp}. We use a machine with 1 GPU and 8 CPUs for our experiments.

\subsection{Results}
For multiple problem scales determined by map size ($\{5 \times 5, 8 \times 8\}$), maximum number of orders ($order \in \{5, 10\}$) and number of pick-up locations in the map ($n \in \{2, 3\}$), we conduct experiments to compare the behavior of RL and the MIP baseline solutions. We examine the trained RL policy's ability to generalize to different order distributions. The hyperparameters used for algorithm training are taken from RLLib robotics examples without fine-tuning. Overall, the RL approach outperforms the baseline across different instance sizes, and generalizes well for unseen order patterns. 

Figure \ref{fig:vrp_mapsize}-\ref{fig:vrp_ordernumber} compares the episodic rewards for the RL policy and the baseline algorithm during training. The shaded band around the mean line shows the minimum and maximum rewards. For readability, the graphs are clipped to skip the initial $3.5$ hours of training as the rewards are highly negative and skew the Y-axis scale. With larger map size or higher order number, the training time required for the agent to achieve rewards equivalent to baseline is higher. This is expected as both the observation and action space increase, the agent requires more exploration to converge to a reasonable policy. Even after three days of training, the rewards for larger instances keep growing gradually. The agent slowly learns to fully utilize the vehicle capacity and to not accept orders which are likely to incur penalty. 

As the agent is trained longer, there is potential for the policy to overfit. In order to test generalizability, we train another policy with a shifted hot order-zone distribution $(0.1, 0.5, 0.3, 0.1)$, and evaluate against the baseline results both using the original order-zone distribution $(0.5, 0.3, 0.1, 0.1)$. Table \ref{table:vrp_baseline_comp} summarizes the evaluation results. It is observed that this policy is able to outperform the baseline consistently during evaluation phase.

We also present the rewards with and without the order miss penalty $q_{failure}$ for the same trained policy to further understand the agent's behavior about order delivery misses. The reward values are close for problems with fewer number of pick-up locations and fewer orders. As the number of pick-up locations becomes larger, the gap between the rewards increases. One explanation is the agent struggles with the increased complexity of order deliveries from different pick-up locations, and its action often changes in the middle of a delivery, so the likelihood of missing the order delivery increases. This behavior is also seen if the number of orders is higher. Even though the RL agent reward is better than the baseline, there is still scope for improvement by reducing the number of order delivery misses.

\begin{table}[h!]
	\resizebox{\columnwidth}{!}{%
		\begin{tabular}{|c|cc|c|}
			\hline
			\multicolumn{1}{|l|}{\multirow{2}{*}{Problem Instance}}                          & \multicolumn{2}{|c|}{RL Evaluation Reward}                                                                               & \multirow{2}{*}{MIP Reward} \\
			\multicolumn{1}{|l|}{}                                                                               & \multicolumn{1}{l}{Without $q_{failure}$ }                                 & \multicolumn{1}{l|}{With $q_{failure}$ }                             &                             \\ \hline
			\begin{tabular}[c]{@{}c}5 by 5 map, 5 orders\\ 2 pick-up locations \end{tabular}                    & \begin{tabular}[c]{@{}c@{}}854.45 \\ (136.03)\end{tabular} & \begin{tabular}[c]{@{}c@{}}838.30 \\ (154.12)\end{tabular} & 595.91                      \\ \hline
			\begin{tabular}[c]{@{}c}5 by 5 map, 5 orders\\ 3 pick-up locations\end{tabular}  & \begin{tabular}[c]{@{}c@{}}754.27 \\ (116.48)\end{tabular} & \begin{tabular}[c]{@{}c@{}}730.40 \\ (132.75)\end{tabular} & 642.62                      \\ \hline
			\begin{tabular}[c]{@{}c}5 by 5 map, 10 orders\\ 2 pick-up locations \end{tabular}               & \begin{tabular}[c]{@{}c@{}}774.63 \\ (143.34)\end{tabular} & \begin{tabular}[c]{@{}c@{}}692.34\\ (200.65)\end{tabular}  & 640.01                      \\ \hline
			\begin{tabular}[c]{@{}c}8 by 8 map, 5 orders\\ 2 pick-up locations\end{tabular}                      & \begin{tabular}[c]{@{}c@{}}548.53 \\ (107.40)\end{tabular} & \begin{tabular}[c]{@{}c@{}}536.55 \\ (112.33)\end{tabular} & 410.58                      \\ \hline
			\begin{tabular}[c]{@{}c}8 by 8 map, 5 orders\\ 3 pick-up locations \end{tabular}                          & \begin{tabular}[c]{@{}c@{}}429.20 \\ (102.37)\end{tabular} & \begin{tabular}[c]{@{}c@{}}373.7 \\ (129.98)\end{tabular}  & 246.25                      \\ \hline
		\end{tabular}
	}
	\caption{RL and baseline solution comparison for VRP. Values in the brackets are standard deviations and mean reward is calculated using 50 episodes.}
	\label{table:vrp_baseline_comp}
\end{table}

\section{Conclusion and Future Work}	
In this paper, we have established Deep Reinforcement Learning (DRL) benchmarks for three canonical dynamic resource allocation problems: Bin Packing, Newsvendor, and Vehicle Routing. We formulated a Markov Decision Process for each problem, and compared established algorithms with vanilla RL techniques. In each case, RL policy either outperforms or is competitive with the baseline. While we do not overcome the NP-hardness of the problems, as wall-clock training time scales with problem size, we find that DRL is a good tool for these problems. 
These results illustrate the potential value of RL for a wide range of real-world industrial online stochastic problems, from order assignment, to retail buying, to real-time routing. Our experiments indicate the following issues as important for making RL solutions more practical in the future: building effective simulators, learning from historical data, initialization of the RL model, overfitting to a particular distribution and enforcement of constraints (e.g. via action masking). 
We used out-of-the-box RL algorithms, with almost no problem-specific tweaking and simple 2-layer neural networks.  Further research can add value by testing various RL algorithms, neural network structures, etc. and seeing their relative value in each problem especially as problem complexity scales up (i.e., solving real-world instances of these problems will likely require innovation on the RL side).  In this paper we only looked at canonical, theoretical models. Further research should endeavor to apply these RL techniques to real-world industrial problems.

\bibliographystyle{aaai}
\bibliography{refs}

\begin{thebibliography}{}

\bibitem[\protect\citeauthoryear{Agrawal and
  Devanur}{2015}]{AgrawalDevanur2015}
Agrawal, S., and Devanur, N.~R.
\newblock 2015.
\newblock Fast algorithms for online stochastic convex programming.
\newblock In {\em Proceedings of the Twenty-sixth Annual ACM-SIAM Symposium on
  Discrete Algorithms}, SODA '15,  1405--1424.
\newblock Philadelphia, PA, USA: Society for Industrial and Applied
  Mathematics.

\bibitem[\protect\citeauthoryear{Andrychowicz \bgroup et al\mbox.\egroup
  }{2018}]{andrychowicz2018learning}
Andrychowicz, M.; Baker, B.; Chociej, M.; Jozefowicz, R.; McGrew, B.; Pachocki,
  J.; Petron, A.; Plappert, M.; Powell, G.; Ray, A.; et~al.
\newblock 2018.
\newblock Learning dexterous in-hand manipulation.
\newblock {\em arXiv preprint arXiv:1808.00177}.

\bibitem[\protect\citeauthoryear{Bello \bgroup et al\mbox.\egroup
  }{2016}]{bello2016neural}
Bello, I.; Pham, H.; Le, Q.~V.; Norouzi, M.; and Bengio, S.
\newblock 2016.
\newblock Neural combinatorial optimization with reinforcement learning.
\newblock {\em arXiv preprint arXiv:1611.09940}.

\bibitem[\protect\citeauthoryear{Brockman \bgroup et al\mbox.\egroup
  }{2016}]{brockman2016openai}
Brockman, G.; Cheung, V.; Pettersson, L.; Schneider, J.; Schulman, J.; Tang,
  J.; and Zaremba, W.
\newblock 2016.
\newblock {OpenAI Gym}.
\newblock {\em arXiv preprint arXiv:1606.01540}.

\bibitem[\protect\citeauthoryear{Coffman~Jr. \bgroup et al\mbox.\egroup
  }{2013}]{Coffmanetal2013}
Coffman~Jr., E.~G.; Csirik, J.; Galambos, G.; Martello, S.; and Vigo, D.
\newblock 2013.
\newblock {\em Bin Packing Approximation Algorithms: Survey and
  Classification}.
\newblock New York, NY: Springer New York.
\newblock  455--531.

\bibitem[\protect\citeauthoryear{Cordeau, Laporte, and
  Ropke}{2008}]{Cordeau2008}
Cordeau, J.-F.; Laporte, G.; and Ropke, S.
\newblock 2008.
\newblock Recent models and algorithms for one-to-one pickup and delivery
  problems.
\newblock In Golden, B.; Raghavan, S.; and Wasil, E., eds., {\em The Vehicle
  Routing Problem: Latest Advances and New Challenges}, volume~43. Boston, MA:
  Springer.

\bibitem[\protect\citeauthoryear{Courcobetis and
  Weber}{1990}]{courcobetis1990stability}
Courcobetis, C., and Weber, R.
\newblock 1990.
\newblock Stability of on-line bin packing with random arrivals and
  long-run-average constraints.
\newblock {\em Probability in the Engineering and Informational Sciences}
  4(4):447--460.

\bibitem[\protect\citeauthoryear{Csirik \bgroup et al\mbox.\egroup
  }{2006}]{csirik2006sum}
Csirik, J.; Johnson, D.~S.; Kenyon, C.; Orlin, J.~B.; Shor, P.~W.; and Weber,
  R.~R.
\newblock 2006.
\newblock On the sum-of-squares algorithm for bin packing.
\newblock {\em Journal of the ACM (JACM)} 53(1):1--65.

\bibitem[\protect\citeauthoryear{Duan \bgroup et al\mbox.\egroup
  }{2016}]{rllab}
Duan, Y.; Chen, X.; Houthooft, R.; Schulman, J.; and Abbeel, P.
\newblock 2016.
\newblock Benchmarking deep reinforcement learning for continuous control.
\newblock {\em CoRR} abs/1604.06778.

\bibitem[\protect\citeauthoryear{Eksioglu, Vural, and
  Reisman}{2009}]{EKSIOGLU20091472}
Eksioglu, B.; Vural, A.~V.; and Reisman, A.
\newblock 2009.
\newblock The vehicle routing problem: A taxonomic review.
\newblock {\em Computers \& Industrial Engineering} 57(4):1472 -- 1483.

\bibitem[\protect\citeauthoryear{Furtado, Munari, and
  Morabito}{2017}]{FURTADO2017334}
Furtado, M. G.~S.; Munari, P.; and Morabito, R.
\newblock 2017.
\newblock Pickup and delivery problem with time windows: A new compact
  two-index formulation.
\newblock {\em Operations Research Letters} 45(4):334 -- 341.

\bibitem[\protect\citeauthoryear{Gijsbrechts \bgroup et al\mbox.\egroup
  }{2018}]{gijsbrechts2018can}
Gijsbrechts, J.; Boute, R.~N.; Van~Mieghem, J.~A.; and Zhang, D.
\newblock 2018.
\newblock Can deep reinforcement learning improve inventory management?
  performance and implementation of dual sourcing-mode problems.
\newblock {\em Performance and Implementation of Dual Sourcing-Mode Problems
  (December 17, 2018)}.

\bibitem[\protect\citeauthoryear{Gupta and Molinaro}{2014}]{GuptaMolinaro2014}
Gupta, A., and Molinaro, M.
\newblock 2014.
\newblock How experts can solve lps online.
\newblock In Schulz, A.~S., and Wagner, D., eds., {\em Algorithms - ESA 2014},
  517--529.
\newblock Berlin, Heidelberg: Springer Berlin Heidelberg.

\bibitem[\protect\citeauthoryear{Gupta and Radovanovic}{2012}]{gupta2012online}
Gupta, V., and Radovanovic, A.
\newblock 2012.
\newblock Online stochastic bin packing.
\newblock {\em arXiv preprint arXiv:1211.2687}.

\bibitem[\protect\citeauthoryear{Horgan \bgroup et al\mbox.\egroup
  }{2018}]{horgan2018distributed}
Horgan, D.; Quan, J.; Budden, D.; Barth-Maron, G.; Hessel, M.; Van~Hasselt, H.;
  and Silver, D.
\newblock 2018.
\newblock Distributed prioritized experience replay.
\newblock {\em arXiv preprint arXiv:1803.00933}.

\bibitem[\protect\citeauthoryear{Huh \bgroup et al\mbox.\egroup
  }{2009}]{huh2009asymptotic}
Huh, W.~T.; Janakiraman, G.; Muckstadt, J.~A.; and Rusmevichientong, P.
\newblock 2009.
\newblock Asymptotic optimality of order-up-to policies in lost sales inventory
  systems.
\newblock {\em Management Science} 55(3):404--420.

\bibitem[\protect\citeauthoryear{Iyengar and Sigman}{2004}]{IyengarSigman2004}
Iyengar, G., and Sigman, K.
\newblock 2004.
\newblock Exponential penalty function control of loss networks.
\newblock {\em Ann. Appl. Probab.} 14(4):1698--1740.

\bibitem[\protect\citeauthoryear{Janakiraman, Seshadri, and
  Shanthikumar}{2007}]{janakiraman2007comparison}
Janakiraman, G.; Seshadri, S.; and Shanthikumar, J.~G.
\newblock 2007.
\newblock A comparison of the optimal costs of two canonical inventory systems.
\newblock {\em Operations Research} 55(5):866--875.

\bibitem[\protect\citeauthoryear{Johnson \bgroup et al\mbox.\egroup
  }{1974}]{Johnsonetal1974}
Johnson, D.; Demers, A.; Ullman, J.; Garey, M.; and Graham, R.
\newblock 1974.
\newblock Worst-case performance bounds for simple one-dimensional packing
  algorithms.
\newblock {\em SIAM Journal on Computing} 3(4):299--325.

\bibitem[\protect\citeauthoryear{Khalil \bgroup et al\mbox.\egroup
  }{2017}]{khalil2017learning}
Khalil, E.; Dai, H.; Zhang, Y.; Dilkina, B.; and Song, L.
\newblock 2017.
\newblock Learning combinatorial optimization algorithms over graphs.
\newblock In {\em Advances in Neural Information Processing Systems},
  6348--6358.

\bibitem[\protect\citeauthoryear{Kong \bgroup et al\mbox.\egroup
  }{2018}]{kong2018new}
Kong, W.; Liaw, C.; Mehta, A.; and Sivakumar, D.
\newblock 2018.
\newblock A new dog learns old tricks: Rl finds classic optimization
  algorithms.

\bibitem[\protect\citeauthoryear{Kool, van Hoof, and
  Welling}{2018}]{kool2018attention}
Kool, W.; van Hoof, H.; and Welling, M.
\newblock 2018.
\newblock Attention, learn to solve routing problems!

\bibitem[\protect\citeauthoryear{Liang \bgroup et al\mbox.\egroup
  }{2018}]{liang2018rllib}
Liang, E.; Liaw, R.; Nishihara, R.; Moritz, P.; Fox, R.; Goldberg, K.;
  Gonzalez, J.~E.; Jordan, M.~I.; and Stoica, I.
\newblock 2018.
\newblock {RLlib}: Abstractions for distributed reinforcement learning.
\newblock In {\em International Conference on Machine Learning ({ICML})}.

\bibitem[\protect\citeauthoryear{Lin \bgroup et al\mbox.\egroup
  }{2018}]{lin2018efficient}
Lin, K.; Zhao, R.; Xu, Z.; and Zhou, J.
\newblock 2018.
\newblock Efficient large-scale fleet management via multi-agent deep
  reinforcement learning.
\newblock In {\em Proceedings of the 24th ACM SIGKDD International Conference
  on Knowledge Discovery \& Data Mining},  1774--1783.
\newblock ACM.

\bibitem[\protect\citeauthoryear{Lu and Dessouky}{2004}]{LuDessouky2004}
Lu, Q., and Dessouky, M.
\newblock 2004.
\newblock An exact algorithm for the multiple vehicle pickup and delivery
  problem.
\newblock {\em Transportation Science} 38(4):503--514.

\bibitem[\protect\citeauthoryear{Mahmoudi and Zhou}{2016}]{MAHMOUDI201619}
Mahmoudi, M., and Zhou, X.
\newblock 2016.
\newblock Finding optimal solutions for vehicle routing problem with pickup and
  delivery services with time windows: A dynamic programming approach based on
  state–space–time network representations.
\newblock {\em Transportation Research Part B: Methodological} 89:19 -- 42.

\bibitem[\protect\citeauthoryear{Nazari \bgroup et al\mbox.\egroup
  }{2018}]{nazari2018reinforcement}
Nazari, M.; Oroojlooy, A.; Snyder, L.; and Tak{\'a}c, M.
\newblock 2018.
\newblock Reinforcement learning for solving the vehicle routing problem.
\newblock In {\em Advances in Neural Information Processing Systems},
  9861--9871.

\bibitem[\protect\citeauthoryear{Oroojlooyjadid \bgroup et al\mbox.\egroup
  }{2017}]{oroojlooyjadid2017deep}
Oroojlooyjadid, A.; Nazari, M.; Snyder, L.; and Tak{\'a}{\v{c}}, M.
\newblock 2017.
\newblock A deep q-network for the beer game: A reinforcement learning
  algorithm to solve inventory optimization problems.
\newblock {\em arXiv preprint arXiv:1708.05924}.

\bibitem[\protect\citeauthoryear{Oroojlooyjadid, Snyder, and
  Tak{\'a}{\v{c}}}{2016}]{oroojlooyjadid2016applying}
Oroojlooyjadid, A.; Snyder, L.; and Tak{\'a}{\v{c}}, M.
\newblock 2016.
\newblock Applying deep learning to the newsvendor problem.
\newblock {\em arXiv preprint arXiv:1607.02177}.

\bibitem[\protect\citeauthoryear{Ritzinger, Puchinger, and
  Hartl}{2016}]{Ulrike2016}
Ritzinger, U.; Puchinger, J.; and Hartl, R.~F.
\newblock 2016.
\newblock A survey on dynamic and stochastic vehicle routing problems.
\newblock {\em International Journal of Production Research} 54(1):215--231.

\bibitem[\protect\citeauthoryear{Ropke and Cordeau}{2009}]{RopkeCordeau2009}
Ropke, S., and Cordeau, J.-F.
\newblock 2009.
\newblock Branch and cut and price for the pickup and delivery problem with
  time windows.
\newblock {\em Transportation Science} 43(3):267--286.

\bibitem[\protect\citeauthoryear{Rudin and Vahn}{2014}]{rudin2014big}
Rudin, C., and Vahn, G.-Y.
\newblock 2014.
\newblock The big data newsvendor: Practical insights from machine learning.

\bibitem[\protect\citeauthoryear{Schulman \bgroup et al\mbox.\egroup
  }{2015}]{schulman2015trust}
Schulman, J.; Levine, S.; Abbeel, P.; Jordan, M.~I.; and Moritz, P.
\newblock 2015.
\newblock Trust region policy optimization.
\newblock In {\em {ICML}}, volume~37,  1889--1897.

\bibitem[\protect\citeauthoryear{Schulman \bgroup et al\mbox.\egroup
  }{2017}]{schulman2017proximal}
Schulman, J.; Wolski, F.; Dhariwal, P.; Radford, A.; and Klimov, O.
\newblock 2017.
\newblock Proximal policy optimization algorithms.
\newblock {\em arXiv preprint arXiv:1707.06347}.

\bibitem[\protect\citeauthoryear{Silver \bgroup et al\mbox.\egroup
  }{2017}]{silver2017mastering}
Silver, D.; Schrittwieser, J.; Simonyan, K.; Antonoglou, I.; Huang, A.; Guez,
  A.; Hubert, T.; Baker, L.; Lai, M.; Bolton, A.; et~al.
\newblock 2017.
\newblock Mastering the game of go without human knowledge.
\newblock {\em Nature} 550(7676):354.

\bibitem[\protect\citeauthoryear{Zipkin}{2000}]{zipkin2000foundations}
Zipkin, P.~H.
\newblock 2000.
\newblock {\em Foundations of inventory management}.
\newblock McGraw-Hill.

\bibitem[\protect\citeauthoryear{Zipkin}{2008}]{zipkin2008old}
Zipkin, P.~H.
\newblock 2008.
\newblock Old and new methods for lost-sales inventory systems.
\newblock {\em Operations Research} 56(5):1256--1263.

\end{thebibliography}

\clearpage

\appendix
\appendixpage

\section{Bin Packing - HyperParameters}
\label{appendix:bin_size_hp}
\begin{table}[h!]
	\resizebox{\columnwidth}{!}
	{
		\centering
		\begin{tabular}{ |c|c|c|c| } 
			\hline
			Discount factor & 0.995 & KL coefficient & 1.0 \\
			\hline
			Experience Buffer & 320000 & Learning rate & 0.0001 \\
			\hline	
			SGD Mini-batch & 32768 & Epochs & 10 \\
			\hline
			Entropy coefficient & 0 & \# Workers & 31 \\
			\hline
			Episode length & 10000 & clip param & 0.3 \\
			\hline
		\end{tabular}
	}
	\caption{Hyperparameters used in PPO for Bin Packing}
	\label{table:bin_packing_hyperparam}
	\vspace{-1em}
\end{table}

\section{Bin Packing Results - Bin Size of 9}
\label{appendix:bin_size_9}

\begin{table}[!htbp]
	\centering
	\begin{tabular}{|c|cc|cc|cc|}
		\hline
		\multicolumn{1}{|l|}{\multirow{2}{*}{Algorithm}}  & \multicolumn{2}{c|}{Perfect Pack} & \multicolumn{2}{c|}{Bounded Waste}	& \multicolumn{2}{c|}{Linear Waste} \\
		\multicolumn{1}{|l|}{} 											&   $\mu$            &        $\sigma$		& 	$\mu$    			& 	$\sigma$     		 &     $\mu$        &      $\sigma$          \\ \hline
		RL with PP  										  			   &  -39.4     		   &          15.1        	&  	-29.0 	  			  &   5.7	 				 &     -106.8       &      70.8     \\ \hline
		RL with BW  									  				  & -57K     		   &          24K        &  	-5.06 	  			  &   2.77	 				 &     -79K       &      8.5K     \\ \hline
		RL with LW 										 				  &  -47.1     		  &         7.7               &  	-41.3	  			&   11.8	 			   &     -71.8      &      10     \\ \hline
		SS  											   						&  -50.2     	    &         28.6           &    -17.27 	  			&   3.21	 			   &     -212.2      &      68.7     \\ \hline
		Best Fit  															  &  -123.7     	   &          8.3           &  	-127.49 	  			&   9.6	 			   &     -130.6     &      7.7     \\ \hline
	\end{tabular}
	\caption{RL and baseline solution comparison for bin packing with bin size of 9 and 1000 items. Mean and standard deviations are calculated across 100 episodes. Note that RL policy trained with BW distribution does not generalize well to PP and LW distribution, giving very negative rewards. Also note that SS outperforms BF for the bin size 9 test case compared to bin size 100.}
	\label{table:bin_size_9_RL_baseline_comp}
\end{table}

\section{Newsvendor - HyperParameters}
\label{appendix:newsvendor_hp}

\begin{table}[h!]
	\centering
	\begin{tabular}{ |c|c| }
		\hline
		Learning rate & 0.00001\\
		\hline
		SGD Mini-Batch Size & 32768\\
		\hline
		Train Batch Size & 320000\\
		\hline
		Episode length & 40\\
		\hline
		Discount factor & 1\\
		\hline
		Epochs & 5\\
		\hline
		Neural Network & Fully connected, 64x32\\
		\hline
		Action Space & Normalized between 0 to 1\\
		\hline
	\end{tabular}
	\caption{Hyperparameters used in PPO for Newsvendor}
	\label{table:newsvendor_hyperparam}
	\vspace{-1em}
\end{table}

\clearpage
\section{VRP baseline MIP formulation}
\label{appendix:vrp_mip}

\subsubsection*{Sets}
\begin{vardefs*}
	V & Current vehicle location, $V=\{0\}$ \\
	P & Pickup nodes (copies of the restaurant nodes, associated with the orders that are not in transit)  \\
	D & Delivery nodes representing the orders that are not in transit, $D = \{j | j= i + n, i \in P, n=|P| \}$  \\
	A & Nodes representing the orders that are accepted by the driver; $A \subset D$ \\
	T & Delivery nodes representing the orders that are in transit  \\
	R & Nodes representing the restaurants, used for final return) \\
	N & Set of all nodes in the graph, $N = V \cup P \cup D  \cup T \cup R $\\
	E & Set of all edges, $E=\{(i, j),  \forall i, j \in N\}$
\end{vardefs*}

\subsubsection*{Decision variables}
\begin{vardefs*}
	x_{ij} & Binary variable, 1 if the vehicle uses the arc from node $i$ to $j$, 0 otherwise; $i, j \in N$ \\
	y_{i}  & Binary variable, 1 if the order $i$ is accepted, 0 otherwise; $i \in P$\\
	Q_{i} & Auxiliary variable to track the capacity usage as of node  $i$; $i \in N$ \\ 
	B_{i} & Auxiliary variable to track the time as of node  $i$; $i \in N$
\end{vardefs*}

\subsubsection*{Parameters}
\begin{vardefs*}
	n & Number of orders available to pick up, $n = |P|$ \\ 
	c_{ij} & Symmetric Manhattan distance (in miles) matrix between node $i$ and $j$, $(i, j) \in E$ \\
	q_i & Supply (demand) at node $i$, $q_0 = |T|; q_i = 1, \forall i \in P;  q_i = -1, \forall i \in D \cup T; q_i = 0 \in R$  \\ 
	l_i & Remaining time to deliver order $i$, $i \in D \cup T$ \\ 
	m & Travel cost per mile \\
	r_i & Revenue for order associated with pick up node $i$, $i \in P$  \\
	U & Vehicle capacity  \\
	M & A very big number  \\ 
	t & Time to travel one mile  \\
	d & A constant positive service time spent on accept, pickup, delivery
\end{vardefs*}

\subsubsection*{Model}
\begin{equation}
\begin{array}{rrclcl}
& \max_{x, y, Q, B} & \multicolumn{3}{l}{ \sum_{i \in P} r_i y_i - m \sum_{(i,j) \in E} c_{ij} x_{ij}  } \\  
& \textrm{s.t.} \qquad  \sum_{j \in N} x_{ij}   &=& y_i & \forall i \in P \\   
& \sum_{j \in N} x_{ij} - \sum_{j \in N} x_{i+n,j}  & = & 0 & \forall i \in P  \\     
& y_i & =& 1 & \forall i \in A \\  
& \sum_{j \in N} x_{ij}   &=& 1 & \forall i \in V \cup T \\  
& \sum_{i \in N \setminus R } \sum_{j \in R }  x_{ij}   &=& 1  &\\     
& \sum_{j \in N \setminus R } x_{ji} - \sum_{j \in N} x_{ij}  & = & 0 & \forall i \in P \cup D \cup T  \\     
& Q_i + q_j - M (1-x_{ij} ) &\leq& Q_j & \forall i,j \in N \\  
& \max{(0, q_i)}    &\leq& Q_i & \forall i \in N \\  
& \min{(U, U+q_i)}    &\geq& Q_i & \forall i \in N \\  
& B_i + d + c_{ij} t -  M (1-x_{ij} )  &\leq& B_j   & \forall i,j \in N \\  
& B_i +  c_{i, i+n} t -  M (1- y_i )  &\leq& B_{i+n}  & \forall i \in P \\ 
&  d \sum_{i \in P \setminus A}  y_i  & = & B_0  \\ 
& B_i  &\leq& l_i  & \forall i \in D \cup T \\ 
& x_{ij}, y_i &\in& \{0, 1\} & \forall i,j \in N  \\ 
\end{array}
\end{equation}

\clearpage

\section{Vehicle Routing Problem - HyperParameters}
\label{appendix:vrp_hp}
\begin{table}[h!]
	\centering
	\begin{tabular}{ |c|c|c|c| } 
		\hline
		Replay buffer alpha & 0.5 & \# steps for Q & 3 \\ 
		\hline
		Replay buffer eps & 0.1 & Learning rate & 1e-3 \\
		\hline	
		Final explore eps & 0.01 & Adam epsilon & 1.5e-4 \\
		\hline
		Replay buffer size & 1e6 & \# Workers & 7 \\
		\hline
		Episode length & 1000 &  Training Batch & 512 \\
		\hline
	\end{tabular}
	\caption{Hyperparameters used in APEX-DQN for VRP, taken directly from the atari APEX example provided in RLLib. No hand-tuning was performed.}
	\label{table:vrp_hyperparam}
	\vspace{-1em}
\end{table}

\end{document}